\documentclass[10pt, a4paper]{article}
\usepackage{lrec2022} 
\usepackage{multibib}
\newcites{languageresource}{Language Resources}
\usepackage{graphicx}
\usepackage{tabularx}
\usepackage{soul}
\usepackage{titlesec}
\titleformat{\section}{\normalfont\large\bfseries\center}{\thesection.}{1em}{}
\titleformat{\subsection}{\normalfont\SmallTitleFont\bfseries\raggedright}{\thesubsection.}{1em}{}
\titleformat{\subsubsection}{\normalfont\normalsize\bfseries\raggedright}{\thesubsubsection.}{1em}{}
\renewcommand\thesection{\arabic{section}}
\renewcommand\thesubsection{\thesection.\arabic{subsection}}
\renewcommand\thesubsubsection{\thesubsection.\arabic{subsubsection}}

\usepackage{epstopdf}
\usepackage[utf8]{inputenc}

\usepackage{hyperref}
\usepackage{xstring}

\usepackage{color}

\title{Trends, Limitations and Open Challenges in \\  Automatic Readability Assessment Research}

\name{Sowmya Vajjala} 

\address{National Research Council, Canada \\
        sowmya.vajjala@nrc-cnrc.gc.ca}

\abstract{
Readability assessment is the task of evaluating the reading difficulty of a given piece of text. This article takes a closer look at contemporary NLP research on developing computational models for readability assessment, identifying the common approaches used for this task, their shortcomings, and some challenges for the future. Where possible, the survey also connects computational research with insights from related work in other disciplines such as education and psychology.    
\\ \newline  }

\begin{document}

\maketitleabstract
\section{Introduction}
\label{sec:intro}
Automatic Readability Assessment (ARA) refers to the task of modeling the reading and comprehension difficulty of a given piece of text, for a given target audience. This has a broad range of applications in both machine facing and human facing scenarios. Some examples of human facing scenarios are: choosing appropriate reading materials for language teaching \cite{Collins-Thompson.Callan-04}, supporting readers with learning disabilities \cite{Rello.Saggion.ea-12} and self-directed learning \cite{Beinborn.Zesch.ea-12}. In machine facing scenarios, ARA is used in scenarios such as for ranking search results by their reading level \cite{Kim.Collins-Thompson.ea-12}, generating translated text controlled for reading level \cite{Marchisio.Guo.ea-19,Agrawal.Carpuat-19}, and evaluating automatic text simplification \cite{Alva-Manchego.Scarton.ea-20}. TextEvaluator \textsuperscript{\texttrademark}\footnote{\url{https://textevaluator.ets.org/TextEvaluator/}}, used to determine whether a reading material is appropriate for a grade level in classroom instruction, is a well known real world application of ARA. Apart from such uses around and within the field of NLP, the general idea of readability assessment is used in a range of other scenarios. A common case is of medical research, where it was used for assessing patient education materials \cite{Sare.Patel.ea-20} and consent forms \cite{Perni.Rooney.ea-19,Lyatoshinsky.Prastsini.ea-19}, for example. This broad application range highlights ARA as one of the important applications of NLP. 

Research into measuring how difficult (or easy) is a text to read is now a century old (e.g., \newcite{Thorndike-21}, \newcite{Lively.Pressey-23}, \newcite{Vogel.Washburne-28}). Early research focused on creating lists of difficult words and/or developing a readability ``formula", which is a simple weighted linear function of easy to calculate variables such as number/length of words/sentences in a text, percentage of difficult words etc. This resulted in several readability formulas such as Flesch Reading Ease \cite{Flesch-48}, SMOG \cite{Mclaughlin-69}, Dale-Chall readability formula \cite{Dale.Chall-48} etc. (see Dubay \shortcite{Dubay-07} for a detailed survey of such formulae).  

NLP researchers started taking interest in this problem only in the past two decades. From statistical language models and feature engineering based machine learning approaches to more recent deep neural networks, a range of approaches have been explored so far for this task. Despite this, a lot of application scenarios involving the use of ARA rely on traditional formulae even within NLP. For example, \newcite{Marchisio.Guo.ea-19} uses the ``traditional formulae" such as Dale-Chall, Flesch Reading ease etc. as a measure of readability to control the reading level of machine translated text. In the scenarios outside of NLP, such as the use cases in medical research mentioned earlier too, one would notice the strong domination of traditional formulae. Possible reasons for this situation could be a lack of awareness of the state of the art in ARA or difficulty in using and interpreting it easily for their purpose.  

Analyzing the reasons for this scenario would require taking a closer look at current methods in ARA research to understand the limitations in its adaptability. To our knowledge, there has only been one comprehensive ARA survey
\cite{Collins-Thompson-14} so far. There have been a lot of newer approaches to ARA since then, and researchers in other disciplines such as education have also published their perspectives on validation and evaluation of ARA approaches (e.g., Hiebert and Pearson \shortcite{Hiebert.Pearson-14}). Further, the approach of the previous survey was also oriented more towards NLP researchers working on ARA. In this background, this paper aims to take a fresh look at ARA considering inputs from other disciplines where needed, and also cover recent research on various aspects of ARA, to get a generalized and contemporary picture about this NLP task.

The paper starts with an overview of the topic (Sections~\ref{sec:intro} and ~\ref{sec:relw}) and summarizes contemporary ARA research in NLP by identifying some common trends (Section ~\ref{sec:trends}). It then discusses their shortcomings (Section~\ref{sec:shortcomings}) in an attempt to understand why this large body of research is not reflected in its usage in various application scenarios. Finally, it identifies some challenges for future research (Section ~\ref{sec:challenges}). Where possible, insights from other disciplines is summarized as well. Note that the terms readability and text complexity are used interchangeably in this paper, as is common in NLP research, although one can see more fine grained difference between the usage of these terms in education or psychology literature (e.g., Valencia et al. \shortcite{Valencia.Wixson.ea-14}). 

This survey is potentially useful for three kinds of readers:
\begin{enumerate}
    \item NLP Researchers specifically working on ARA and other related problems (e.g., text simplification) may find this survey useful to understand the task holistically and identify  language specific challenges. 
     \item Other NLP researchers can get a general overview of ARA and how to incorporate it into their systems. 
    \item Researchers from other disciplines looking to use ARA for their research can get an overview of the state of research in the field and what they can use easily. 
\end{enumerate}

\section{Related Work}
\label{sec:relw}
While there was been a lot of work in the NLP community on developing computational models for readability assessment across languages, there has not been much work synthesizing this research. \newcite{Collins-Thompson-14} is the most recent, comprehensive survey on this topic, to our knowledge. It gave a detailed overview of the various approaches to ARA and identified the development of user-centric models, data driven measures that can be easily specialized to new domains, and the inclusion of domain/conceptual knowledge into existing models as some of the potential research directions for future. 
\newcite{Francois-15} presented a historical overview of readability assessment focusing on early research on traditional formulae and identified three challenges for future work - validity of the training data, developing ARA approaches for different domains, and difficulty in estimating readability at different granularities (e.g., words and sentences). 

Outside of NLP, \newcite{Nelson.Perfetti.ea-12} compared and evaluated a few existing proprietary text difficulty metrics (for English) using a range of reading difficulty annotated corpora and assessed the implications of such measures for education. In 2014, the Elementary School Journal published a special issue on understanding text complexity \cite{Hiebert.Pearson-14}, which offered multi-disciplinary perspectives on various aspects of ARA and its implications to education. Concluding that readability involves dimensions other than text and much more research is needed on the topic, the special issue cautioned about the danger of focusing on text readability scores alone. While the last two are not survey articles per se, they are included here as they summarize the findings from research that is not common knowledge in NLP research on ARA.   

In the current survey, the focus is more on the recent developments in ARA research in NLP, drawing inputs from existing body of research in other related disciplines as needed. The goal of this paper is to provide a general overview of the trends in research and not to have an exhaustive listing of all published research on this topic during this period. The paper aims to remain language agnostic in this study, focusing primarily on the approaches taken for corpus creation, modeling and evaluation.  

\section{Current Trends in ARA Research in NLP}
\label{sec:trends}
\begin{center}
\begin{figure}[htb!]
    \centering
    \includegraphics[width=0.5\textwidth]{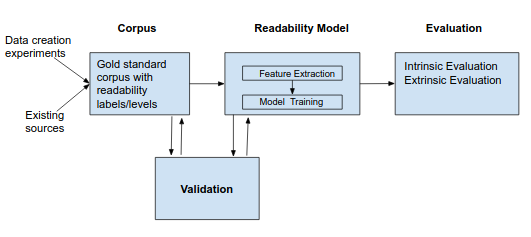}
    \caption{Typical ARA pipeline}
    \label{fig:arapipeline}
\end{figure}
\end{center}

ARA is generally modeled as a supervised machine learning problem in NLP literature. Hence, a typical ARA approach follows the pipeline depicted in Figure~\ref{fig:arapipeline}.  

ARA approaches rely on a gold standard training corpus annotated with labels indicating reading level categories, or numbers indicating a graded scale (\textbf{Corpus}). As with any machine learning problem, the next step consists of feature extraction and training a model (\textbf{Readability Model}). The final step in this process is an evaluation of the effectiveness of the model (\textbf{Evaluation}). A not so commonly seen, but essential step in this process is \textbf{Validation}, which evaluates not the model, but the process itself, including the corpora and features. Rest of this section discusses each of these steps in detail by giving an overview of representative approaches taken by researchers in handling these stages of ARA, and what changed in the recent few years, compared to \newcite{Collins-Thompson-14}'s survey.

\subsection{Corpus}
\label{sec:corpora}
\begin{figure}
    \centering
    \includegraphics[width=0.45\textwidth]{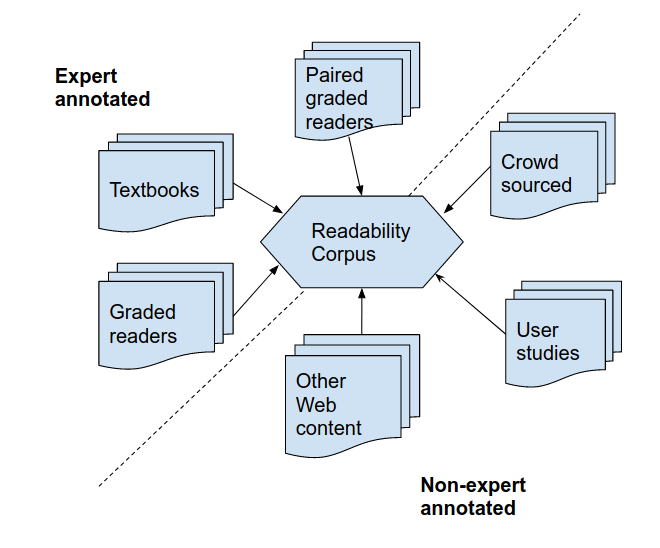}
    \caption{Various forms of ARA Corpora}
    \label{fig:aracorpora}
\end{figure}

Training data in ARA comes from various sources. They can be broadly classified into two categories: expert annotated and non-expert annotated. Textbooks, or other graded readers carefully prepared by trained authors targeting audience at specific grade levels can be termed as ``expert annotated". These are the most common forms of training data seen in ARA research. On the other hand, some ARA work also relied on available web content, or doing crowd sourcing experiments and user studies to collect data. In such cases, we either do not know who did the annotations or we are getting them from a target reader population, who need not be experts on the linguistic aspects of readability. Figure ~\ref{fig:aracorpora} summarizes the different forms of data sources in ARA research, and the rest of this section discusses each of them in detail.

\paragraph{Textbooks: } Textbooks have been a common source of training data for ARA research, when available, for several languages such as English \cite{Heilman.Collins-Thompson.ea-07}, Japanese \cite{Sato.Matsuyoshi.ea-08}, German \cite{Berendes.Vajjala.ea-18}, Swedish \cite{Pilan.Vajjala.ea-16}, French \cite{Francois.Fairon-12}, Bangla \cite{Islam.Mehler.ea-12} and Greek \cite{Chatzipanagiotidis.Giagkov.ea-21}, to name a few. They are considered to be naturally suited for ARA research as one would expect the linguistic characteristics of texts to become more complex as school grade increases. On a related note, \newcite{Xia.Kochmar.ea-16} collected reading comprehension passages from language exams conducted at different proficiency levels for building ARA models. 

However, it is not always possible to have a readily accessible dataset of textbooks, as many textbooks are also under copyright and may not be accessible in a digitized form. Thus, most of the above mentioned corpora are not available for other researchers, which makes them a valuable, but not viable data source. A closer alternative is to use graded readers. 

\paragraph{Graded Readers: } This paper refers to non-textbook reading materials prepared by teachers or other experts, which are separated into some categorization of reading levels, as graded readers. Typically, these materials are derived from sources such as: news articles rewritten to suit the target reading level, encyclopedia articles written for adults and children separately, or children's readers from book publishing companies. WeeBit  \cite{Vajjala.Meurers-12} is one of the widely used graded reader corpus used for English ARA. Such graded readers exist for other languages as well. For example, \newcite{Imperial-21} recently described one such dataset for Filipino language. 

In the recent past, corpora such as Newsela \cite{Xu.Callison-Burch.ea-15} and OneStopEnglish \cite{Vajjala.Lucic-18} were created for English, which can be called \textbf{Paired Graded Readers}. Instead of having a collection of unrelated documents at each reading level, these corpora have the same documents rewritten to suit different reading levels. Newsela corpus, which also has a Spanish subset, was used to build text simplification systems \cite{Stajner.Nisioi-18} and generating machine translated text at varying reading levels \cite{Agrawal.Carpuat-19} in the past. 

\paragraph{Other web content: } When there are no available texts annotated with reading level, it is common to find other documents from the web which have some form of inherent reading level grouping. Simple Wikipedia\footnote{\url{https://simple.wikipedia.org/}} was widely used along with Wikipedia to build a easy versus difficult ARA system for English \cite{Napoles.Dredze-10}. A sentence aligned version of this dataset was also used for automatic text simplification \cite{Hwang.Hajishirzi.ea-15}. 

Other such websites have been used in other ARA approaches for English \cite{Vajjala.Meurers-13}, German \cite{Hancke.Vajjala-12}, Italian \cite{DellOrletta.Montemagni.ea-11} and Basque \cite{Gonzalez.Aranzabe.ea-14} among others. \cite{Azpiazu.Pera-19} used Vikidia\footnote{\url{https://www.vikidia.org/}} together with Wikipedia to compile a multilingual readability dataset in 7 languages.

Taking a slightly different approach, \newcite{Eickhoff.Serdyukov.ea-11} relied on the topic hierarchy in Open Directory Project to group web articles based on whether they are appropriate to a certain age group. \newcite{Vajjala.Meurers-14a} used  a corpus of TV program subtitles grouped into three age groups, collected from BBC channels. In the absence of readily available corpora annotated with reading levels, this seems to be the most common way of procuring some form of leveled text corpus for this task. 

\paragraph{Crowdsourcing: } All the above mentioned approaches relied on some form of an existing data source suitable for training ARA models. \newcite{Declercq.Hoste.ea-14} described the usefulness of a crowdsourcing for ARA, where non-expert readers/general public are shown two unrelated texts (in Dutch) each time and are asked to compare them in terms of their reading difficulty. Comparing these judgments with expert (e.g., teacher) judgments, they concluded that crowdsourcing is a viable alternative to expert annotations for this task. 

\paragraph{User studies: } Another way to gather an ARA corpus is by conducting user studies. For example, \newcite{Bruck.Hartrumpf.ea-08} conducted a user study with 500 German documents from municipal domain, and non-expert readers were asked to rate the texts on a 7 point Likert scale \cite{Likert-32} and used it to construct an ARA model. Similarly, \newcite{Pitler.Nenkova-08} conducted a user study where college students were asked to rate WSJ news articles on a scale, which was then used to build a readability model.

\newcite{Stajner.Ponzetto.ea-17} collected user judgements of sentence level text complexity in the context of text simplification, for original, manually and automatically simplified sentences. Some studies conducted such studies to gather expert annotations as well. For example, \newcite{Kate.Luo.ea-10} described a dataset collected through a user study, rated separately by experts and naive readers. \newcite{Shen.Williams.ea-13} used a dataset collected and annotated by experts, in four languages - Arabic, Dari, English, and Pashto. Note that this is different from using available textbooks or graded readers, which are also graded by experts. User studies are typically conducted specifically for this task, and not for a generic use as in the case of other expert annotated resources.  

Eye tracking and reading time information were also used in the past to annotate readability datasets \cite{Nishikawa.Makino.ea-13,Yaneva.Temnikova.ea-15}, which were done with less number of participants than the other user studies mentioned above. However, overall, user studies are not a common mode of corpus creation for ARA, owing to the time and effort involved. They also typically  result in smaller datasets compared to other approaches for this task. 

Among these different forms of resources, excepting paired graded readers and very few cases from "other web content", the texts/vocabulary at different reading levels in the corpora don't necessarily deal with the same content. For example, in the WeeBit corpus \cite{Vajjala.Meurers-12}, one of the commonly used corpus for English, articles tagged with different reading levels don't share the same topical content. As we will see in the next subsection, a majority of ARA models do not particularly control for topic variation. This leads us to question what the ARA models learn - is it a notion of text complexity, or topical differences among texts? 

Further, whether these corpora are validated to be appropriate for the target audience is another important concern, not typically addressed in ARA research. For example, Simple Wikipedia is written for children and adults learning English. However, there is no evidence in the form of a user study that shows that this is indeed the case. Yet, it is used, along with Wikipedia, as a common data source for building readability models. Recently, \newcite{Vajjala.Lucic-19}'s study with over 100 participants concluded that the reading level annotations for texts in a paired graded corpus did not have any effect on reader's comprehension. In this background, an obvious question that arises is - what is the right corpus for this problem? \newcite{Francois-15} too discussed the issue of validity of training data in the context of ARA and called for more work in this direction. 

Another potential problem with existing ARA datasets is that of inter-annotator agreement. With user studies and crowd sourcing based data collection, it is possible to gather such information. However, we have no means of acquiring this information for other texts, especially the expert annotated corpora. It could be hard to understand and identify the shortcomings of ARA approaches without having a clear picture of human agreement on the task. 

Finally, an often ignored issue in the discussion around ARA datasets is the domain of the texts. Textbooks and  news articles seem to be the most commonly used genre, although we see focused datasets on ARA for legal/government documents, literary pieces etc. There is some past research that looked into genre effect on ARA models \cite{Nelson.Perfetti.ea-12,Dellorletta.Montemagni.ea-14} and on how to develop an unbiased model across genres \cite{Sheehan.Flor.ea-13}. However, this is an essential, but under-explored aspect in ARA research so far.

To conclude, while there are many ways of creating corpora for ARA research, we don't have many freely available corpora covering different languages, topics, and target domains, and we don't have strongly validated corpora suited for this task. Compared to \newcite{Collins-Thompson-14}'s survey, we can say that not much has happened in terms of corpora creation in ARA, and many questions remain.

\subsection{Readability Model}
\label{sec:modeling}

The second step in ARA pipeline is to build the readability model, which includes both the feature extraction/text representation as well as training an ARA model. Research into building readability models in the past two decades has primarily relied on language models and feature engineering based machine learning approaches. Like with other NLP tasks, recent approaches relied on neural network and deep learning approaches for this task. 

Features that are expected to influence the readability of a text come in various forms, from some simple, easy to calculate numbers such as number of words per sentence to more complex ones involving the estimation of a discourse structure in the document. While some of the advanced linguistic features such as coherence and cohesion are potentially hard to extract automatically, shallow variants e.g., noun overlap between adjacent sentences, implemented in Coh-Metrix \cite{Graesser.McNamara.ea-04} are commonly used as proxies. Similarly, different kinds of text embeddings, which capture some form of syntactic and semantic properties of texts, also do not need advanced linguistic processing such as parsing, coreference resolution etc. Hence, instead of grouping features based on linguistic categories, as is commonly done, they are grouped based on the amount of language processing required in this paper. 

\begin{figure*}
    \centering
    \includegraphics[width=0.8\textwidth]{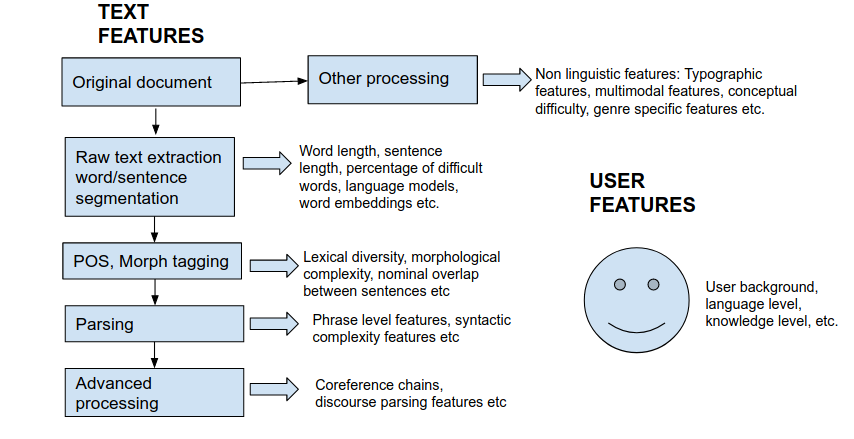}
    \caption{Features used in ARA grouped by the amount of language processing needed}
    \label{fig:arafeatures}
\end{figure*}

Figure ~\ref{fig:arafeatures} shows a summary of different kinds of features used in ARA research with examples at each step.

\paragraph{Feature Engineering: } 
Features such as word length (in characters/syllables), sentence length, usage of different forms of word lists \cite{Chen.Meurers-18}, language models (e.g., \newcite{Petersen-09}), models of word acquisition \cite{Kidwell.Lebanon.ea-11}, measures of morphological variation and complexity \cite{Hancke.Vajjala-12,Chatzipanagiotidis.Giagkov.ea-21}, syntactic complexity \cite{Heilman.Collins-Thompson.ea-07,Vajjala.Meurers-12}, psycholinguistic processes \cite{Howcroft.Demberg-17} and other attributes have been extensively used for developing ARA models across languages. Some features relying on advanced processing such as coreference resolution and discourse relations \cite{Pitler.Nenkova-08,Feng.Jansche.ea-10} have also been explored in the past, more for English, and to some extent for other languages such as French \cite{Todirascu.Francois.ea-16}. \cite{Collins-Thompson-14} presents a comprehensive summary of different kinds of features used in ARA.

 Some recent research focused on learning task specific embeddings (e.g., \newcite{Cha.Gwon.ea-17}, \newcite{Jiang.Gu.ea-18}) for ARA. Although not common, there has also been some work on modeling conceptual difficulty \cite{Jameel.Lam.ea-12}. An often ignored aspect of ARA is the reader. \newcite{Kim.Collins-Thompson.ea-12} is one of rare works related to ARA which considers reader attributes such as interests, language level etc. into their model to rank search results by their reading level. Although not directly about ARA, \newcite{Knowles.Renduchintala.ea-16} explored the relationship between a word comprehension and a learner's native language. More recently, \newcite{Gooding.Berzak.ea-21} proposed a method to predict text readability from the reader's scrolling behavior. Overall, though ARA approaches are meant to be for real users in most of the cases, we don't see much work on modeling user features in relation to ARA. 
 
 Feature engineering based ARA approaches typically employ feature selection methods to choose a subset of features that best work for the task from a larger set. Apart from generic methods such as information gain, feature correlation etc., genetic algorithm based optimization methods were also explored for this task \cite{De.Hoste-16}. Although some papers report on "best performing features" for a given dataset, we don't have a clear consensus on what groups of features perform better across languages and dataset. In a recent work, \newcite{Weiss.Chen.ea-21} showed experimented with English and German ARA using a broad linguistic feature set and presented a study of what features are consistently useful for both languages, and what are not, for this task. More research in this direction is needed to gain a better understanding of a core set of useful linguistic features for ARA across languages.
 
 \paragraph{Training: }
 In terms of training methods used, ARA is generally modeled as a supervised learning problem, especially classification. It is, however, not uncommon to see it being modeled as regression \cite{Vajjala.Meurers-14b} and ranking \cite{Tanaka-10,Ma.Fosler-Lussier.ea-12,Lee.Vajjala-22}. \newcite{Heilman.Collins-Thompson.ea-08} compared different approaches to learn an ARA model and showed that ordinal regression is better suited for the task. \newcite{Xia.Kochmar.ea-16} showed that pair wise ranking approach may generalize better compared to classification. Unlike such approaches, \newcite{Jiang.Gu.ea-19} proposed a graph propagation based approach to ARA, which can potentially consider the inter-relationships between documents while modeling readability. Finally, while almost all of ARA research has been modeling it as a supervised learning problem, \newcite{Martinc.Pollak.ea-21} and \newcite{Ehara-21} proposed unsupervised approaches to measuring text readability in the recent past.

Like other NLP research, ARA in the past two years has been dominated by neural network based architectures. For example, \newcite{Mohammadi.Khasteh-19} proposed a multilingual readability assessment model using deep reinforcement learning and \newcite{Meng.Chen.ea-20} proposed ReadNet, a hierarchical self attention based transformer model for ARA. Contemporary research also explored different ways of combining linguistic features with transformer models \cite{Deutsch.Jasbi.ea-20,Lee.Jang.ea-21}. 

In general, most readability approaches have been shown to work for one language, or individual models were developed for each language. However, Azpiazu and Pera  \shortcite{Azpiazu.Pera-19,Azpiazu.Pera-20} study the development of multilingual and cross-lingual approaches to ARA using deep learning architectures. \newcite{Weiss.Chen.ea-21} studied whether a common core of linguistic features would be useful across languages, and performed zero-shot cross-lingual evaluation between English and German using a large collection of linguistic features.

To summarize, we may notice  that the past two decades of ARA research closely followed other areas of NLP i.e, traditional feature engineering based methods heavily dominated most of the previous research, whereas recent research seems to see more deep learning based approaches. Compared to the previous survey from 2014, most new research on ARA seems to have focused particularly on this aspect. Yet, there doesn't seem to be a clear consensus on what works for ARA across languages. While \newcite{Lee.Jang.ea-21} concluded that a combination of transformer architecture and linguistic features give a better performance, \newcite{Weiss.Chen.ea-21} showed zero shot cross lingual transfer with linguistic features alone. More recently, \newcite{Lee.Vajjala-22} proposed a neural pairwise ranking model, that showed good zero shot cross-lingual transfer with only BERT embeddings as the starting point. So, while deep learning has clearly been useful for ARA, linguistic features still seem to show strong results for languages with existing NLP tools such as POS taggers and syntactic parsers. 

\subsection{Evaluation}
\label{sec:eval}
Evaluation of ARA can happen in two forms: intrinsic (evaluating on a standard test set) and extrinsic (evaluating on an end task). Most of the papers describing ARA models evaluate them intrinsically in terms of classification accuracy, F-score, Pearson/Spearman correlation (regression/ranking approaches), root mean square error (regression) and other such measures on held-out test data or in a cross-validated setup, as is conventionally done while evaluating supervised machine learning approaches. While it is not a default, we also see multi-corpus evaluation e.g., training on one corpus, testing on many; training and testing on many corpora \cite{Nelson.Perfetti.ea-12,Vajjala.Meurers-14b,Xia.Kochmar.ea-16}. Another way of evaluating if texts predicted by a ARA model to be ``simple" result in better comprehension for the target reader group is through a user study. To our knowledge, such an evaluation has not been conducted so far for ARA models.

In terms of extrinsic evaluation, \newcite{Pera.Ng-12} and \newcite{Kim.Collins-Thompson.ea-12} reported on experiments related to integrating readability approach into a search engine, and applying it for personalized search. \newcite{Sheehan.Kostin.ea-14} deployed ARA models into a real-world tool. However, these examples are more of exceptions than norms, and such extrinsic evaluation is rare in ARA research, perhaps owing to the time and effort involved in such endeavours. 

\subsection{Validation}
Validation is the step of assessing the accuracy of a process. Validation is distinct from evaluation as we are here evaluating other stages in model building, and not the ARA model itself. In the context of ARA research, validation is the step that answers the following two questions:
\begin{enumerate}\itemsep-1ex
    \item Are the reading level differences annotated in text corpora actually reflected in a reader's experience with the texts? i.e., Does the (annotated) reading level have any relation to reader comprehension? 
    \item Are the features used to represent a text theoretically valid, and can they reliably learn the reading level differences among texts?
\end{enumerate}

Although these questions seem obvious, and have been posed many times in non-computational work on text readability in the past (e.g., \newcite{Cunningham.Mesmer-14}), there is not much work in this direction in contemporary ARA research in NLP. Research related to TextEvaluator \cite{Sheehan.Kostin.ea-14,Sheehan-17} has the only detailed analysis in this direction, to our knowledge. However, these are published outside of typical NLP venues, and hence, may not draw the attention of ARA researchers within NLP. 

\newcite{Francois-14} conducted a qualitative and quantitative analysis of a French as Foreign Language textbook corpus and concluded that there is a lack of consistent correlation among expert ratings, and that the texts assigned at the same level by the expert annotators showed significant differences in terms of lexical and syntactic features. \newcite{Berendes.Vajjala.ea-18} reached similar conclusions using a multidimensional corpus of graded German textbooks covering two school tracks and four publishers. 

Although there are a few user studies aiming to study the relationship between readability annotations and reader comprehension \cite{Crossley.Yang.ea-14,Vajjala.Meurers.ea-16,Vajjala.Lucic-19}, conclusions have been mixed. The most recent among these, \newcite{Vajjala.Lucic-19}'s study concluded that the reading level annotations for texts in a paired graded corpus did not have any effect on reader's comprehension. 

To summarize, validation is an essential step in understanding whether our ARA models are really capturing the notion of text complexity, or just modeling randomly captured patterns in a given dataset. Clearly, there is not much work done on validation in ARA research, and this is an area which needs further work. Now that we know about the trends in ARA research at different stages of building and evaluating a model, what is lacking?

 \section{Limitations}
 \label{sec:shortcomings}
 Based on this overview of current trends in the corpora creation, modeling, evaluation and validation of ARA, I identify the following limitations that are potentially preventing the adaption of modern ARA techniques into other research areas within and outside NLP.
 
 \begin{enumerate}\itemsep-0.5ex
     \item \textbf{Multidimensional and Multimodal ARA models: } - Text readability involves several aspects of text, starting from typographical to linguistic, from conceptual difficulty to deeper pragmatics. However, contemporary ARA research tends to focus on the surface textual form. Topical or conceptual difficulty is not given much importance. Where it is considered, it is typically not combined with other aspects of readability. 
     
     Further, texts don't exist in isolation. Many times, there is accompanying non-text data such as tables and/or images in the document. Although psycholinguists and cognitive psychologists explored such aspects through eye tracking studies in the past, I am are not aware of any research that touches upon these aspects in the context of NLP. To summarize, there is no framework yet (to our knowledge) that can incorporate a multidimensional, multimodal view of text complexity. 

     \item \textbf{Reader and Task considerations: } Research in education and psychology typically describes text complexity as a combination of text properties, reader (user) characteristics, and task complexity \cite{Goldman.Lee-14,Valencia.Wixson.ea-14}. However, within NLP, ARA research is almost always focused on text, with a small amount of research on reader modeling \cite{Kim.Collins-Thompson.ea-12,Gooding.Berzak.ea-21} and how what is complex can depend on a reader's language proficiency \cite{Gooding.Kochmar.ea-21}. While some research on modeling task complexity started to emerge \cite{Kuhberger.Bramann.ea-19}, I am not aware of any approach that considers task complexity in the context of ARA or combine all the three aspects.
     
          \item \textbf{Availability of corpus resources: } There is clearly a lot of work on ARA across many languages. Yet, we don't don't see a lot of publicly available corpora. Even when available, one has ask whether the corpora suit the target scenario. For example, one cannot use a corpus of textbooks to evaluate ARA models that intend to serve, say, dyslexic readers, as the reading difficulties experienced by dyslexic readers are completely different from first language readers learning subject matter in school. Similarly, it is not appropriate to use a corpus of news articles to develop a readability measure for legal texts. Such lack of available (and diverse) corpora can limit the development of ARA models tailored to specific application scenarios.
          
         \item \textbf{Availability of ready to use tools: } There is not much of readily usable code artefacts related to building and using ARA models online. While some researchers shared code to reproduce their experiments (e.g., \newcite{Ambati.Reddy.ea-16}, \newcite{Howcroft.Demberg-17}), there is not much usable code for other NLP researchers or off the shelf tools for researchers from other disciplines. Recent tools such as LingFeat \cite{Lee.Jang.ea-21} provide implementations to a wide range of linguistic features, including traditional readability formulae, but don't have any ready to use pre-trained readability systems. Availability of such tools can potentially be useful for researchers from other disciplines wanting to use readability assessment approaches to answer research questions in their own domains.     
         \item \textbf{Lack of extrinsic evaluation: } Typically, ARA approaches are evaluated intrinsically, using cross validation or held out test set. It is rare to see an extrinsic evaluation when we consider a typical ARA research paper. This makes it particularly hard for practitioners to understand whether an approach works in an applied scenario. 
     \item \textbf{Lack of validation and interpretation: } The most common approach taken in building an ARA model is to take an available corpus, extract various kinds of features, and train different models and compare them. However, there is very little research on whether the corpus is suited for the task, whether the features themselves are actually useful, or if they have a theoretical grounding. Further, it is hard to understand what exactly does a model learn about text complexity. These issues make it difficult for researchers from other domains wanting to adapt modern ARA methods, and they instead turn to traditional formulae, which are relatively straight forward to interpret, even if they themselves are not validated either. 
     
 \end{enumerate}
 
 Although some of these limitations can be termed generic to NLP itself and not specific to ARA, this section attempted to highlight these issues in the context of contemporary ARA approaches. Among these, the first three limitations are of particular concern to NLP researchers, both in terms of using ARA in other NLP problems as well as furthering research on ARA itself. The remaining limitations are more general in nature, and would interest all the three target audience. I believe these are the factors that come between ARA research and its broader usefulness.
 
 
 \section{Challenges and Open Questions}
 \label{sec:challenges}
 
 In view of the above mentioned limitations and their potential consequences, I identify four major challenge areas where more future work is needed to address the current limitations of ARA.
 
 \begin{enumerate}\itemsep-0.5ex
 \item \textbf{A framework to develop a holistic model of text readability}: We have seen that ARA research is primarily focused on textual features, especially those that focus on form. However, there are many other aspects such as conceptual difficulty, typographic features, user characteristics, task features etc, as we saw earlier. An obvious challenge would be to develop a unified model of ARA that encompasses all these aspects. However, it is not the work of one person or group, nor can it all be done in one go. So, an important first step in this direction (which can address limitations 1--2) would be to design an easily extendable framework to build a holistic model of readability by incrementally adding multiple dimensions, covering multi modal data. This would also necessitate the development of appropriate corpora and other resources suitable for this purpose. 
 
 \item \textbf{Models adaptable to new domain}: Any ARA model could still only be relevant to the target domain/audience and may not directly transfer to a new application scenario. Hence, approaches that can transfer an existing model into a new domain/audience should be developed. One potential avenue to explore in this direction is to model ARA as a ranking problem instead of classification or regression, as recent research concludes that it generalizes better than other models \cite{Lee.Vajjala-22}. This can address the limitation 3 mentioned earlier. 
 
  \item \textbf{Creation of open and diverse datasets and tools: } Development of openly accessible corpora which suit various application scenarios, for several languages is a major challenge in ARA research, as we saw earlier. New methods to quickly create (and validate) corpora need to be developed. Whether recent developments in data augmentation can be useful for developing ARA corpora is also something that can be explored in future. For widespread adaptation of research on ARA, and to progress towards a holistic model, ready to use tools should be developed. Tools such as Coh-Metrix \cite{Graesser.McNamara.ea-11} and  LingFeat\footnote{\url{https://github.com/brucewlee/lingfeat}} \cite{Lee.Jang.ea-21} that provide a range of linguistic features typically associated with readability assessment are a step in this direction. Along with these, tools that can show the predictions of ARA models should also be developed, to address the limitations 3--4. 

 \item \textbf{Developing Best Practices: } To support the creation of reusable resources (corpora/code) and to be able to reproduce/replicate results and understand SOTA, a set of best practices must be developed for ARA. Some inspiration for this can be drawn from the procedures and findings of the recently conducted REPROLANG challenge \cite{Branco.Calzolari.ea-20} which conducted a shared task to replicate some published NLP research. The best practices for ARA should also include guidelines for validating the corpora and features developed, as well as recommended procedures for developing interpretable approaches. This can help one address the limitations 5--6 to some extent. This will also potentially encourage non-NLP researchers to seriously consider employing more recent ARA models in their research. Some aspects of this challenge area (e.g., validation, interpretation) demand expertise beyond NLP methods and may require inter-disciplinary collaborations. 
 \end{enumerate}
 It has to be noted that some of these challenges are not necessarily specific to ARA, and are applicable across NLP in general. However, as with the previous section, this paper aims to discuss them in the context of ARA in particular and not in the context of entire NLP research. Further, This collection of ideas on challenges for future is by no means exhaustive, and I hope this survey initiates more discussion in this direction. 

\section{Conclusion}
\label{sec:concl}
In this paper, I presented an overview of two decades of research on automatic readability assessment in NLP and connected it with related areas of research and applications. During this process I identified the limitations of contemporary research and identified some challenge areas for future. This analysis leads us to conclude that despite a large body of research, we don't yet have a clear picture of what are a good set of resources, modeling techniques that can be considered as SOTA across languges in ARA. There is also a dearth of off the shelf tools and resources that support researchers and practitioners interested in ARA. Further, many challenges mentioned in previous surveys still remain. Considering that readability assessment has a wide range of applications in and outside NLP as it was seen from examples in Section~\ref{sec:intro}, I think it is important to address these issues and enable the a broader adaption of ARA approaches within and outside NLP, over traditional formulae which only consider superficial aspects of language. More focus on validating NLP approaches to ARA, and on being able to interpret and relate model predictions to actual textual complexity may be the first steps in this direction. 

\section*{Acknowledgements}
I thank the anonymous reviewers for their detailed comments on the submitted version, Michel Simard and Rebecca Knowles for their comments on an earlier version of the paper, and Joseph Mervin Imperial, for actionable feedback before the final submission. 

\section{Bibliographical References}\label{reference}

\bibliographystyle{lrec2022-bib}
\bibliography{lrec2022}

\begin{thebibliography}{}

\bibitem[\protect\citename{Agrawal and Carpuat}2019]{Agrawal.Carpuat-19}
Agrawal, S. and Carpuat, M.
\newblock (2019).
\newblock Controlling text complexity in neural machine translation.
\newblock In {\em Proceedings of the 2019 Conference on Empirical Methods in
  Natural Language Processing and the 9th International Joint Conference on
  Natural Language Processing (EMNLP-IJCNLP)}, pages 1549--1564.

\bibitem[\protect\citename{Alva-Manchego \bgroup et al.\egroup
  }2020]{Alva-Manchego.Scarton.ea-20}
Alva-Manchego, F., Scarton, C., and Specia, L.
\newblock (2020).
\newblock Data-driven sentence simplification: Survey and benchmark.
\newblock {\em Computational Linguistics}, 46(1):135--187.

\bibitem[\protect\citename{Ambati \bgroup et al.\egroup
  }2016]{Ambati.Reddy.ea-16}
Ambati, B.~R., Reddy, S., and Steedman, M.
\newblock (2016).
\newblock Assessing relative sentence complexity using an incremental ccg
  parser.
\newblock In {\em Proceedings of the 2016 Conference of the North American
  Chapter of the Association for Computational Linguistics: Human Language
  Technologies}, pages 1051--1057.

\bibitem[\protect\citename{Azpiazu and Pera}2019]{Azpiazu.Pera-19}
Azpiazu, I.~M. and Pera, M.~S.
\newblock (2019).
\newblock Multiattentive recurrent neural network architecture for multilingual
  readability assessment.
\newblock {\em Transactions of the Association for Computational Linguistics},
  7:421--436.

\bibitem[\protect\citename{Azpiazu and Pera}2020]{Azpiazu.Pera-20}
Azpiazu, I.~M. and Pera, M.~S.
\newblock (2020).
\newblock Is cross-lingual readability assessment possible?
\newblock {\em Journal of the Association for Information Science and
  Technology}, 71(6):644--656.

\bibitem[\protect\citename{Beinborn \bgroup et al.\egroup
  }2012]{Beinborn.Zesch.ea-12}
Beinborn, L., Zesch, T., and Gurevych, I.
\newblock (2012).
\newblock Towards fine-grained readability measures for self-directed language
  learning.
\newblock In {\em Proceedings of the SLTC 2012 workshop on NLP for CALL; Lund;
  25th October; 2012}, pages 11--19. Link{\"o}ping University Electronic Press.

\bibitem[\protect\citename{Berendes \bgroup et al.\egroup
  }2018]{Berendes.Vajjala.ea-18}
Berendes, K., Vajjala, S., Meurers, D., Bryant, D., Wagner, W., Chinkina, M.,
  and Trautwein, U.
\newblock (2018).
\newblock Reading demands in secondary school: Does the linguistic complexity
  of textbooks increase with grade level and the academic orientation of the
  school track?
\newblock {\em Journal of Educational Psychology}, 110(4):518.

\bibitem[\protect\citename{Branco \bgroup et al.\egroup
  }2020]{Branco.Calzolari.ea-20}
Branco, A., Calzolari, N., Vossen, P., van Noord, G., Van~Uytvanck, D., Silva,
  J., Gomes, L., Moreira, A., and Elbers, W.
\newblock (2020).
\newblock A shared task of a new, collaborative type to foster reproducibility:
  A first exercise in the area of language science and technology with
  reprolang2020.
\newblock In {\em Proceedings of The 12th Language Resources and Evaluation
  Conference}, pages 5539--5545.

\bibitem[\protect\citename{Cha \bgroup et al.\egroup }2017]{Cha.Gwon.ea-17}
Cha, M., Gwon, Y., and Kung, H.
\newblock (2017).
\newblock Language modeling by clustering with word embeddings for text
  readability assessment.
\newblock In {\em Proceedings of the 2017 ACM on Conference on Information and
  Knowledge Management}, pages 2003--2006.

\bibitem[\protect\citename{Chatzipanagiotidis \bgroup et al.\egroup
  }2021]{Chatzipanagiotidis.Giagkov.ea-21}
Chatzipanagiotidis, S., Giagkou, M., and Meurers, D.
\newblock (2021).
\newblock Broad linguistic complexity analysis for greek readability
  classification.
\newblock In {\em Proceedings of the 16th Workshop on Innovative Use of NLP for
  Building Educational Applications}, pages 48--58.

\bibitem[\protect\citename{Chen and Meurers}2018]{Chen.Meurers-18}
Chen, X. and Meurers, D.
\newblock (2018).
\newblock Word frequency and readability: Predicting the text-level readability
  with a lexical-level attribute.
\newblock {\em Journal of Research in Reading}, 41(3):486--510.

\bibitem[\protect\citename{Collins-Thompson and
  Callan}2004]{Collins-Thompson.Callan-04}
Collins-Thompson, K. and Callan, J.
\newblock (2004).
\newblock Information retrieval for language tutoring: An overview of the reap
  project.
\newblock In {\em Proceedings of the 27th annual international ACM SIGIR
  conference on Research and development in information retrieval}, pages
  544--545.

\bibitem[\protect\citename{Collins-Thompson}2014]{Collins-Thompson-14}
Collins-Thompson, K.
\newblock (2014).
\newblock Computational assessment of text readability: A survey of current and
  future research.
\newblock {\em ITL-International Journal of Applied Linguistics},
  165(2):97--135.

\bibitem[\protect\citename{Crossley \bgroup et al.\egroup
  }2014]{Crossley.Yang.ea-14}
Crossley, S.~A., Yang, H.~S., and McNamara, D.~S.
\newblock (2014).
\newblock What's so simple about simplified texts? a computational and
  psycholinguistic investigation of text comprehension and text processing.
\newblock {\em Reading in a Foreign Language}, 26(1):92--113.

\bibitem[\protect\citename{Cunningham and
  Anne~Mesmer}2014]{Cunningham.Mesmer-14}
Cunningham, J.~W. and Anne~Mesmer, H.
\newblock (2014).
\newblock Quantitative measurement of text difficulty: What’s the use?
\newblock {\em The Elementary School Journal}, 115(2):255--269.

\bibitem[\protect\citename{Dale and Chall}1948]{Dale.Chall-48}
Dale, E. and Chall, J.~S.
\newblock (1948).
\newblock A formula for predicting readability: Instructions.
\newblock {\em Educational research bulletin}, pages 37--54.

\bibitem[\protect\citename{De~Clercq and Hoste}2016]{De.Hoste-16}
De~Clercq, O. and Hoste, V.
\newblock (2016).
\newblock All mixed up? finding the optimal feature set for general readability
  prediction and its application to english and dutch.
\newblock {\em Computational Linguistics}, 42(3):457--490.

\bibitem[\protect\citename{De~Clercq \bgroup et al.\egroup
  }2014]{Declercq.Hoste.ea-14}
De~Clercq, O., Hoste, V., Desmet, B., Van~Oosten, P., De~Cock, M., and Macken,
  L.
\newblock (2014).
\newblock Using the crowd for readability prediction.
\newblock {\em Natural Language Engineering}, 20(3):293--325.

\bibitem[\protect\citename{Dell'Orletta \bgroup et al.\egroup
  }2011]{DellOrletta.Montemagni.ea-11}
Dell'Orletta, F., Montemagni, S., and Venturi, G.
\newblock (2011).
\newblock Read-it: Assessing readability of italian texts with a view to text
  simplification.
\newblock In {\em Proceedings of the second workshop on speech and language
  processing for assistive technologies}, pages 73--83. Association for
  Computational Linguistics.

\bibitem[\protect\citename{Dell’Orletta \bgroup et al.\egroup
  }2014]{Dellorletta.Montemagni.ea-14}
Dell’Orletta, F., Montemagni, S., and Venturi, G.
\newblock (2014).
\newblock Assessing document and sentence readability in less resourced
  languages and across textual genres.
\newblock {\em ITL-International Journal of Applied Linguistics},
  165(2):163--193.

\bibitem[\protect\citename{Deutsch \bgroup et al.\egroup
  }2020]{Deutsch.Jasbi.ea-20}
Deutsch, T., Jasbi, M., and Shieber, S.
\newblock (2020).
\newblock Linguistic features for readability assessment.
\newblock {\em arXiv preprint arXiv:2006.00377}.

\bibitem[\protect\citename{DuBay}2007]{Dubay-07}
DuBay, W.~H.
\newblock (2007).
\newblock {\em Unlocking language: The classic readability studies}.
\newblock Impact Information.

\bibitem[\protect\citename{Ehara}2021]{Ehara-21}
Ehara, Y.
\newblock (2021).
\newblock Evaluation of unsupervised automatic readability assessors using rank
  correlations.
\newblock In {\em Proceedings of the 2nd Workshop on Evaluation and Comparison
  of NLP Systems}, pages 62--72.

\bibitem[\protect\citename{Eickhoff \bgroup et al.\egroup
  }2011]{Eickhoff.Serdyukov.ea-11}
Eickhoff, C., Serdyukov, P., and De~Vries, A.~P.
\newblock (2011).
\newblock A combined topical/non-topical approach to identifying web sites for
  children.
\newblock In {\em Proceedings of the fourth ACM international conference on Web
  search and data mining}, pages 505--514.

\bibitem[\protect\citename{Feng \bgroup et al.\egroup
  }2010]{Feng.Jansche.ea-10}
Feng, L., Jansche, M., Huenerfauth, M., and Elhadad, N.
\newblock (2010).
\newblock A comparison of features for automatic readability assessment.
\newblock In {\em Coling 2010: Posters}, pages 276--284.

\bibitem[\protect\citename{Flesch}1948]{Flesch-48}
Flesch, R.
\newblock (1948).
\newblock A new readability yardstick.
\newblock {\em Journal of applied psychology}, 32(3):221.

\bibitem[\protect\citename{Fran{\c{c}}ois and Fairon}2012]{Francois.Fairon-12}
Fran{\c{c}}ois, T. and Fairon, C.
\newblock (2012).
\newblock An {``}{AI} readability{''} formula for {F}rench as a foreign
  language.
\newblock In {\em Proceedings of the 2012 Joint Conference on Empirical Methods
  in Natural Language Processing and Computational Natural Language Learning},
  pages 466--477, Jeju Island, Korea, July. Association for Computational
  Linguistics.

\bibitem[\protect\citename{Fran{\c{c}}ois}2014]{Francois-14}
Fran{\c{c}}ois, T.
\newblock (2014).
\newblock An analysis of a french as a foreign language corpus for readability
  assessment.
\newblock In {\em Proceedings of the third workshop on NLP for
  computer-assisted language learning}, pages 13--32.

\bibitem[\protect\citename{Fran{\c{c}}ois}2015]{Francois-15}
Fran{\c{c}}ois, T.
\newblock (2015).
\newblock When readability meets computational linguistics: a new paradigm in
  readability.
\newblock {\em Revue fran{\c{c}}aise de linguistique appliqu{\'e}e},
  20(2):79--97.

\bibitem[\protect\citename{Goldman and Lee}2014]{Goldman.Lee-14}
Goldman, S.~R. and Lee, C.~D.
\newblock (2014).
\newblock Text complexity: State of the art and the conundrums it raises.
\newblock {\em the elementary school journal}, 115(2):290--300.

\bibitem[\protect\citename{Gonzalez-Dios \bgroup et al.\egroup
  }2014]{Gonzalez.Aranzabe.ea-14}
Gonzalez-Dios, I., Aranzabe, M.~J., de~Ilarraza, A.~D., and Salaberri, H.
\newblock (2014).
\newblock Simple or complex? assessing the readability of basque texts.
\newblock In {\em Proceedings of COLING 2014, the 25th international conference
  on computational linguistics: Technical papers}, pages 334--344.

\bibitem[\protect\citename{Gooding \bgroup et al.\egroup
  }2021a]{Gooding.Berzak.ea-21}
Gooding, S., Berzak, Y., Mak, T., and Sharifi, M.
\newblock (2021a).
\newblock Predicting text readability from scrolling interactions.
\newblock In {\em Proceedings of the 25th Conference on Computational Natural
  Language Learning}, pages 380--390, Online, November. Association for
  Computational Linguistics.

\bibitem[\protect\citename{Gooding \bgroup et al.\egroup
  }2021b]{Gooding.Kochmar.ea-21}
Gooding, S., Kochmar, E., Yimam, S.~M., and Biemann, C.
\newblock (2021b).
\newblock Word complexity is in the eye of the beholder.
\newblock In {\em Proceedings of the 2021 Conference of the North American
  Chapter of the Association for Computational Linguistics: Human Language
  Technologies}, pages 4439--4449, Online, June. Association for Computational
  Linguistics.

\bibitem[\protect\citename{Graesser \bgroup et al.\egroup
  }2004]{Graesser.McNamara.ea-04}
Graesser, A.~C., McNamara, D.~S., Louwerse, M.~M., and Cai, Z.
\newblock (2004).
\newblock Coh-metrix: Analysis of text on cohesion and language.
\newblock {\em Behavior research methods, instruments, \& computers},
  36(2):193--202.

\bibitem[\protect\citename{Graesser \bgroup et al.\egroup
  }2011]{Graesser.McNamara.ea-11}
Graesser, A.~C., McNamara, D.~S., and Kulikowich, J.~M.
\newblock (2011).
\newblock Coh-metrix: Providing multilevel analyses of text characteristics.
\newblock {\em Educational researcher}, 40(5):223--234.

\bibitem[\protect\citename{Hancke \bgroup et al.\egroup
  }2012]{Hancke.Vajjala-12}
Hancke, J., Vajjala, S., and Meurers, D.
\newblock (2012).
\newblock Readability classification for german using lexical, syntactic, and
  morphological features.
\newblock In {\em Proceedings of COLING 2012}, pages 1063--1080.

\bibitem[\protect\citename{Heilman \bgroup et al.\egroup
  }2007]{Heilman.Collins-Thompson.ea-07}
Heilman, M., Collins-Thompson, K., Callan, J., and Eskenazi, M.
\newblock (2007).
\newblock Combining lexical and grammatical features to improve readability
  measures for first and second language texts.
\newblock In {\em Human Language Technologies 2007: The Conference of the North
  American Chapter of the Association for Computational Linguistics;
  Proceedings of the Main Conference}, pages 460--467.

\bibitem[\protect\citename{Heilman \bgroup et al.\egroup
  }2008]{Heilman.Collins-Thompson.ea-08}
Heilman, M., Collins-Thompson, K., and Eskenazi, M.
\newblock (2008).
\newblock An analysis of statistical models and features for reading difficulty
  prediction.
\newblock In {\em Proceedings of the third workshop on innovative use of NLP
  for building educational applications}, pages 71--79.

\bibitem[\protect\citename{Hiebert and Pearson}2014]{Hiebert.Pearson-14}
Hiebert, E.~H. and Pearson, P.~D.
\newblock (2014).
\newblock Understanding text complexity: Introduction to the special issue.
\newblock {\em the elementary school journal}, 115(2):153--160.

\bibitem[\protect\citename{Howcroft and Demberg}2017]{Howcroft.Demberg-17}
Howcroft, D.~M. and Demberg, V.
\newblock (2017).
\newblock Psycholinguistic models of sentence processing improve sentence
  readability ranking.
\newblock In {\em Proceedings of the 15th Conference of the European Chapter of
  the Association for Computational Linguistics: Volume 1, Long Papers}, pages
  958--968.

\bibitem[\protect\citename{Hwang \bgroup et al.\egroup
  }2015]{Hwang.Hajishirzi.ea-15}
Hwang, W., Hajishirzi, H., Ostendorf, M., and Wu, W.
\newblock (2015).
\newblock Aligning sentences from standard wikipedia to simple wikipedia.
\newblock In {\em Proceedings of the 2015 Conference of the North American
  Chapter of the Association for Computational Linguistics: Human Language
  Technologies}, pages 211--217.

\bibitem[\protect\citename{Imperial}2021]{Imperial-21}
Imperial, J.~M.
\newblock (2021).
\newblock Bert embeddings for automatic readability assessment.
\newblock In {\em Proceedings of the International Conference on Recent
  Advances in Natural Language Processing (RANLP 2021)}, pages 611--618.

\bibitem[\protect\citename{Islam \bgroup et al.\egroup
  }2012]{Islam.Mehler.ea-12}
Islam, Z., Mehler, A., and Rahman, R.
\newblock (2012).
\newblock Text readability classification of textbooks of a low-resource
  language.
\newblock In {\em Proceedings of the 26th Pacific Asia Conference on Language,
  Information, and Computation}, pages 545--553.

\bibitem[\protect\citename{Jameel \bgroup et al.\egroup
  }2012]{Jameel.Lam.ea-12}
Jameel, S., Lam, W., and Qian, X.
\newblock (2012).
\newblock Ranking text documents based on conceptual difficulty using term
  embedding and sequential discourse cohesion.
\newblock In {\em 2012 IEEE/WIC/ACM International Conferences on Web
  Intelligence and Intelligent Agent Technology}, volume~1, pages 145--152.
  IEEE.

\bibitem[\protect\citename{Jiang \bgroup et al.\egroup }2018]{Jiang.Gu.ea-18}
Jiang, Z., Gu, Q., Yin, Y., and Chen, D.
\newblock (2018).
\newblock Enriching word embeddings with domain knowledge for readability
  assessment.
\newblock In {\em Proceedings of the 27th International Conference on
  Computational Linguistics}, pages 366--378.

\bibitem[\protect\citename{Jiang \bgroup et al.\egroup }2019]{Jiang.Gu.ea-19}
Jiang, Z., Gu, Q., Yin, Y., Wang, J., and Chen, D.
\newblock (2019).
\newblock Graw+: A two-view graph propagation method with word coupling for
  readability assessment.
\newblock {\em Journal of the Association for Information Science and
  Technology}, 70(5):433--447.

\bibitem[\protect\citename{Kate \bgroup et al.\egroup }2010]{Kate.Luo.ea-10}
Kate, R.~J., Luo, X., Patwardhan, S., Franz, M., Florian, R., Mooney, R.~J.,
  Roukos, S., and Welty, C.
\newblock (2010).
\newblock Learning to predict readability using diverse linguistic features.
\newblock In {\em Proceedings of the 23rd international conference on
  computational linguistics}, pages 546--554. Association for Computational
  Linguistics.

\bibitem[\protect\citename{Kidwell \bgroup et al.\egroup
  }2011]{Kidwell.Lebanon.ea-11}
Kidwell, P., Lebanon, G., and Collins-Thompson, K.
\newblock (2011).
\newblock Statistical estimation of word acquisition with application to
  readability prediction.
\newblock {\em Journal of the American Statistical Association},
  106(493):21--30.

\bibitem[\protect\citename{Kim \bgroup et al.\egroup
  }2012]{Kim.Collins-Thompson.ea-12}
Kim, J.~Y., Collins-Thompson, K., Bennett, P.~N., and Dumais, S.~T.
\newblock (2012).
\newblock Characterizing web content, user interests, and search behavior by
  reading level and topic.
\newblock In {\em Proceedings of the fifth ACM international conference on Web
  search and data mining}, pages 213--222.

\bibitem[\protect\citename{Knowles \bgroup et al.\egroup
  }2016]{Knowles.Renduchintala.ea-16}
Knowles, R., Renduchintala, A., Koehn, P., and Eisner, J.
\newblock (2016).
\newblock Analyzing learner understanding of novel l2 vocabulary.
\newblock In {\em Proceedings of The 20th SIGNLL Conference on Computational
  Natural Language Learning}, pages 126--135.

\bibitem[\protect\citename{K{\"u}hberger \bgroup et al.\egroup
  }2019]{Kuhberger.Bramann.ea-19}
K{\"u}hberger, C., Bramann, C., Wei{\ss}, Z., and Meurers, D.
\newblock (2019).
\newblock Task complexity in history textbooks: A multidisciplinary case study
  on triangulation in history education research.
\newblock {\em History Education Research Journal}, 16(1):139--157.

\bibitem[\protect\citename{Lee and Vajjala}2022]{Lee.Vajjala-22}
Lee, J. and Vajjala, S.
\newblock (2022).
\newblock A neural pairwise ranking model for readability assessment.
\newblock {\em Findings of the Association for Computational Linguistics: ACL
  2022}, May.

\bibitem[\protect\citename{Lee \bgroup et al.\egroup }2021]{Lee.Jang.ea-21}
Lee, B.~W., Jang, Y.~S., and Lee, J. H.-J.
\newblock (2021).
\newblock Pushing on text readability assessment: A transformer meets
  handcrafted linguistic features.
\newblock {\em arXiv preprint arXiv:2109.12258}.

\bibitem[\protect\citename{Likert}1932]{Likert-32}
Likert, R.
\newblock (1932).
\newblock A technique for the measurement of attitudes.
\newblock {\em Archives of psychology}.

\bibitem[\protect\citename{Lively and Pressey}1923]{Lively.Pressey-23}
Lively, B.~A. and Pressey, S.~L.
\newblock (1923).
\newblock A method for measuring the vocabulary burden of textbooks.
\newblock {\em Educational administration and supervision}, 9(389-398):73.

\bibitem[\protect\citename{Lyatoshinsky \bgroup et al.\egroup
  }2019]{Lyatoshinsky.Prastsini.ea-19}
Lyatoshinsky, P., Pratsinis, M., Abt, D., Schmid, H.-P., Zumstein, V., and
  Betschart, P.
\newblock (2019).
\newblock Readability assessment of commonly used german urological
  questionnaires.
\newblock {\em Current urology}, 13(2):87--93.

\bibitem[\protect\citename{Ma \bgroup et al.\egroup
  }2012]{Ma.Fosler-Lussier.ea-12}
Ma, Y., Fosler-Lussier, E., and Lofthus, R.
\newblock (2012).
\newblock Ranking-based readability assessment for early primary children’s
  literature.
\newblock In {\em Proceedings of the 2012 Conference of the North American
  Chapter of the Association for Computational Linguistics: Human Language
  Technologies}, pages 548--552.

\bibitem[\protect\citename{Marchisio \bgroup et al.\egroup
  }2019]{Marchisio.Guo.ea-19}
Marchisio, K., Guo, J., Lai, C.-I., and Koehn, P.
\newblock (2019).
\newblock Controlling the reading level of machine translation output.
\newblock In {\em Proceedings of Machine Translation Summit XVII Volume 1:
  Research Track}, pages 193--203.

\bibitem[\protect\citename{Martinc \bgroup et al.\egroup
  }2021]{Martinc.Pollak.ea-21}
Martinc, M., Pollak, S., and Robnik-{\v{S}}ikonja, M.
\newblock (2021).
\newblock Supervised and unsupervised neural approaches to text readability.
\newblock {\em Computational Linguistics}, 47(1):141--179.

\bibitem[\protect\citename{McLaughlin}1969]{Mclaughlin-69}
McLaughlin, G.~H.
\newblock (1969).
\newblock Smog grading-a new readability formula.
\newblock {\em Journal of reading}, 12(8):639--646.

\bibitem[\protect\citename{Meng \bgroup et al.\egroup }2020]{Meng.Chen.ea-20}
Meng, C., Chen, M., Mao, J., and Neville, J.
\newblock (2020).
\newblock Readnet: A hierarchical transformer framework for web article
  readability analysis.
\newblock In {\em European Conference on Information Retrieval}, pages 33--49.
  Springer.

\bibitem[\protect\citename{Mohammadi and Khasteh}2019]{Mohammadi.Khasteh-19}
Mohammadi, H. and Khasteh, S.~H.
\newblock (2019).
\newblock Text as environment: A deep reinforcement learning text readability
  assessment model.
\newblock {\em arXiv preprint arXiv:1912.05957}.

\bibitem[\protect\citename{Napoles and Dredze}2010]{Napoles.Dredze-10}
Napoles, C. and Dredze, M.
\newblock (2010).
\newblock Learning simple wikipedia: A cogitation in ascertaining abecedarian
  language.
\newblock In {\em Proceedings of the NAACL HLT 2010 Workshop on Computational
  Linguistics and Writing: Writing Processes and Authoring Aids}, pages 42--50.
  Association for Computational Linguistics.

\bibitem[\protect\citename{Nelson \bgroup et al.\egroup
  }2012]{Nelson.Perfetti.ea-12}
Nelson, J., Perfetti, C., Liben, D., and Liben, M.
\newblock (2012).
\newblock Measures of text difficulty: Testing their predictive value for grade
  levels and student performance.
\newblock {\em Council of Chief State School Officers, Washington, DC}.

\bibitem[\protect\citename{Nishikawa \bgroup et al.\egroup
  }2013]{Nishikawa.Makino.ea-13}
Nishikawa, H., Makino, T., and Matsuo, Y.
\newblock (2013).
\newblock A pilot study of readability prediction with reading time.
\newblock In {\em Proceedings of the Second Workshop on Predicting and
  Improving Text Readability for Target Reader Populations}, pages 78--84,
  Sofia, Bulgaria, August. Association for Computational Linguistics.

\bibitem[\protect\citename{Pera and Ng}2012]{Pera.Ng-12}
Pera, M.~S. and Ng, Y.-K.
\newblock (2012).
\newblock Brek12: a book recommender for k-12 users.
\newblock In {\em Proceedings of the 35th international ACM SIGIR conference on
  Research and development in information retrieval}, pages 1037--1038.

\bibitem[\protect\citename{Perni \bgroup et al.\egroup
  }2019]{Perni.Rooney.ea-19}
Perni, S., Rooney, M.~K., Horowitz, D.~P., Golden, D.~W., McCall, A.~R.,
  Einstein, A.~J., and Jagsi, R.
\newblock (2019).
\newblock Assessment of use, specificity, and readability of written clinical
  informed consent forms for patients with cancer undergoing radiotherapy.
\newblock {\em JAMA oncology}, 5(8):e190260--e190260.

\bibitem[\protect\citename{Petersen and Ostendorf}2009]{Petersen-09}
Petersen, S.~E. and Ostendorf, M.
\newblock (2009).
\newblock A machine learning approach to reading level assessment.
\newblock {\em Computer speech \& language}, 23(1):89--106.

\bibitem[\protect\citename{Pil{\'a}n \bgroup et al.\egroup
  }2016]{Pilan.Vajjala.ea-16}
Pil{\'a}n, I., Vajjala, S., and Volodina, E.
\newblock (2016).
\newblock A readable read: Automatic assessment of language learning materials
  based on linguistic complexity.
\newblock {\em arXiv preprint arXiv:1603.08868}.

\bibitem[\protect\citename{Pitler and Nenkova}2008]{Pitler.Nenkova-08}
Pitler, E. and Nenkova, A.
\newblock (2008).
\newblock Revisiting readability: A unified framework for predicting text
  quality.
\newblock In {\em Proceedings of the 2008 conference on empirical methods in
  natural language processing}, pages 186--195.

\bibitem[\protect\citename{Rello \bgroup et al.\egroup
  }2012]{Rello.Saggion.ea-12}
Rello, L., Saggion, H., Baeza-Yates, R., and Graells, E.
\newblock (2012).
\newblock Graphical schemes may improve readability but not understandability
  for people with dyslexia.
\newblock In {\em Proceedings of the First Workshop on Predicting and Improving
  Text Readability for target reader populations}, pages 25--32. Association
  for Computational Linguistics.

\bibitem[\protect\citename{Sare \bgroup et al.\egroup }2020]{Sare.Patel.ea-20}
Sare, A., Patel, A., Kothari, P., Kumar, A., Patel, N., and Shukla, P.~A.
\newblock (2020).
\newblock Readability assessment of internet-based patient education materials
  related to treatment options for benign prostatic hyperplasia.
\newblock {\em Academic Radiology}.

\bibitem[\protect\citename{Sato \bgroup et al.\egroup
  }2008]{Sato.Matsuyoshi.ea-08}
Sato, S., Matsuyoshi, S., and Kondoh, Y.
\newblock (2008).
\newblock Automatic assessment of japanese text readability based on a textbook
  corpus.
\newblock In {\em LREC}.

\bibitem[\protect\citename{Sheehan \bgroup et al.\egroup
  }2013]{Sheehan.Flor.ea-13}
Sheehan, K.~M., Flor, M., and Napolitano, D.
\newblock (2013).
\newblock A two-stage approach for generating unbiased estimates of text
  complexity.
\newblock In {\em Proceedings of the Workshop on Natural Language Processing
  for Improving Textual Accessibility}, pages 49--58.

\bibitem[\protect\citename{Sheehan \bgroup et al.\egroup
  }2014]{Sheehan.Kostin.ea-14}
Sheehan, K.~M., Kostin, I., Napolitano, D., and Flor, M.
\newblock (2014).
\newblock The textevaluator tool: Helping teachers and test developers select
  texts for use in instruction and assessment.
\newblock {\em The Elementary School Journal}, 115(2):184--209.

\bibitem[\protect\citename{Sheehan}2017]{Sheehan-17}
Sheehan, K.~M.
\newblock (2017).
\newblock Validating automated measures of text complexity.
\newblock {\em Educational Measurement: Issues and Practice}, 36(4):35--43.

\bibitem[\protect\citename{Shen \bgroup et al.\egroup
  }2013]{Shen.Williams.ea-13}
Shen, W., Williams, J., Marius, T., and Salesky, E.
\newblock (2013).
\newblock A language-independent approach to automatic text difficulty
  assessment for second-language learners.
\newblock In {\em Proceedings of the Second Workshop on Predicting and
  Improving Text Readability for Target Reader Populations}, pages 30--38.

\bibitem[\protect\citename{{\v{S}}tajner and Nisioi}2018]{Stajner.Nisioi-18}
{\v{S}}tajner, S. and Nisioi, S.
\newblock (2018).
\newblock A detailed evaluation of neural sequence-to-sequence models for
  in-domain and cross-domain text simplification.
\newblock In {\em Proceedings of the eleventh international conference on
  language resources and evaluation (LREC 2018)}.

\bibitem[\protect\citename{{\v{S}}tajner \bgroup et al.\egroup
  }2017]{Stajner.Ponzetto.ea-17}
{\v{S}}tajner, S., Ponzetto, S.~P., and Stuckenschmidt, H.
\newblock (2017).
\newblock Automatic assessment of absolute sentence complexity.
\newblock In {\em Proceedings of the 26th International Joint Conference on
  Artificial Intelligence, IJCAI}, volume~17, pages 4096--4102.

\bibitem[\protect\citename{Tanaka-Ishii \bgroup et al.\egroup }2010]{Tanaka-10}
Tanaka-Ishii, K., Tezuka, S., and Terada, H.
\newblock (2010).
\newblock Sorting texts by readability.
\newblock {\em Computational linguistics}, 36(2):203--227.

\bibitem[\protect\citename{Thorndike}1921]{Thorndike-21}
Thorndike, E.~L.
\newblock (1921).
\newblock {\em The teacher’s word book}.
\newblock Teacher's College, Columbia University.

\bibitem[\protect\citename{Todirascu \bgroup et al.\egroup
  }2016]{Todirascu.Francois.ea-16}
Todirascu, A., Fran{\c{c}}ois, T., Bernhard, D., Gala, N., and Ligozat, A.-L.
\newblock (2016).
\newblock Are cohesive features relevant for text readability evaluation?
\newblock In {\em 26th International Conference on Computational Linguistics
  (COLING 2016)}, pages 987--997.

\bibitem[\protect\citename{Vajjala and Lu{\v{c}}i{\'c}}2018]{Vajjala.Lucic-18}
Vajjala, S. and Lu{\v{c}}i{\'c}, I.
\newblock (2018).
\newblock Onestopenglish corpus: A new corpus for automatic readability
  assessment and text simplification.
\newblock In {\em Proceedings of the thirteenth workshop on innovative use of
  NLP for building educational applications}, pages 297--304.

\bibitem[\protect\citename{Vajjala and Lucic}2019]{Vajjala.Lucic-19}
Vajjala, S. and Lucic, I.
\newblock (2019).
\newblock On understanding the relation between expert annotations of text
  readability and target reader comprehension.
\newblock In {\em Proceedings of the Fourteenth Workshop on Innovative Use of
  NLP for Building Educational Applications}, pages 349--359.

\bibitem[\protect\citename{Vajjala and Meurers}2012]{Vajjala.Meurers-12}
Vajjala, S. and Meurers, D.
\newblock (2012).
\newblock On improving the accuracy of readability classification using
  insights from second language acquisition.
\newblock In {\em Proceedings of the seventh workshop on building educational
  applications using NLP}, pages 163--173. Association for Computational
  Linguistics.

\bibitem[\protect\citename{Vajjala and Meurers}2013]{Vajjala.Meurers-13}
Vajjala, S. and Meurers, D.
\newblock (2013).
\newblock On the applicability of readability models to web texts.
\newblock In {\em Proceedings of the Second Workshop on Predicting and
  Improving Text Readability for Target Reader Populations}, pages 59--68.

\bibitem[\protect\citename{Vajjala and Meurers}2014a]{Vajjala.Meurers-14a}
Vajjala, S. and Meurers, D.
\newblock (2014a).
\newblock Exploring measures of “readability” for spoken language:
  Analyzing linguistic features of subtitles to identify age-specific tv
  programs.
\newblock In {\em Proceedings of the 3rd Workshop on Predicting and Improving
  Text Readability for Target Reader Populations (PITR)}, pages 21--29.

\bibitem[\protect\citename{Vajjala and Meurers}2014b]{Vajjala.Meurers-14b}
Vajjala, S. and Meurers, D.
\newblock (2014b).
\newblock Readability assessment for text simplification: From analysing
  documents to identifying sentential simplifications.
\newblock {\em ITL-International Journal of Applied Linguistics},
  165(2):194--222.

\bibitem[\protect\citename{Vajjala \bgroup et al.\egroup
  }2016]{Vajjala.Meurers.ea-16}
Vajjala, S., Meurers, D., Eitel, A., and Scheiter, K.
\newblock (2016).
\newblock Towards grounding computational linguistic approaches to readability:
  Modeling reader-text interaction for easy and difficult texts.
\newblock In {\em Proceedings of the Workshop on Computational Linguistics for
  Linguistic Complexity (CL4LC)}, pages 38--48.

\bibitem[\protect\citename{Valencia \bgroup et al.\egroup
  }2014]{Valencia.Wixson.ea-14}
Valencia, S.~W., Wixson, K.~K., and Pearson, P.~D.
\newblock (2014).
\newblock Putting text complexity in context: Refocusing on comprehension of
  complex text.
\newblock {\em The Elementary School Journal}, 115(2):270--289.

\bibitem[\protect\citename{Vogel and Washburne}1928]{Vogel.Washburne-28}
Vogel, M. and Washburne, C.
\newblock (1928).
\newblock An objective method of determining grade placement of children's
  reading material.
\newblock {\em The Elementary School Journal}, 28(5):373--381.

\bibitem[\protect\citename{vor~der Br{\"u}ck \bgroup et al.\egroup
  }2008]{Bruck.Hartrumpf.ea-08}
vor~der Br{\"u}ck, T., Hartrumpf, S., and Helbig, H.
\newblock (2008).
\newblock A readability checker with supervised learning using deep indicators.
\newblock {\em Informatica}, 32(4).

\bibitem[\protect\citename{Weiss \bgroup et al.\egroup }2021]{Weiss.Chen.ea-21}
Weiss, Z., Chen, X., and Meurers, D.
\newblock (2021).
\newblock Using broad linguistic complexity modeling for cross-lingual
  readability assessment.
\newblock In {\em Proceedings of the 10th Workshop on NLP for Computer Assisted
  Language Learning}, pages 38--54.

\bibitem[\protect\citename{Xia \bgroup et al.\egroup }2016]{Xia.Kochmar.ea-16}
Xia, M., Kochmar, E., and Briscoe, T.
\newblock (2016).
\newblock Text readability assessment for second language learners.
\newblock In {\em Proceedings of the 11th Workshop on Innovative Use of NLP for
  Building Educational Applications}, pages 12--22.

\bibitem[\protect\citename{Xu \bgroup et al.\egroup
  }2015]{Xu.Callison-Burch.ea-15}
Xu, W., Callison-Burch, C., and Napoles, C.
\newblock (2015).
\newblock Problems in current text simplification research: New data can help.
\newblock {\em Transactions of the Association for Computational Linguistics},
  3:283--297.

\bibitem[\protect\citename{Yaneva \bgroup et al.\egroup
  }2015]{Yaneva.Temnikova.ea-15}
Yaneva, V., Temnikova, I., and Mitkov, R.
\newblock (2015).
\newblock Accessible texts for autism: An eye-tracking study.
\newblock In {\em Proceedings of the 17th International ACM SIGACCESS
  Conference on Computers \& Accessibility}, pages 49--57.

\end{thebibliography}


\bibliographystylelanguageresource{lrec2022-bib}

\end{document}